\documentclass[10pt,twocolumn]{article}

\usepackage{wacv}
\usepackage{times}
\usepackage{epsfig}
\usepackage{graphicx}
\usepackage{amsmath}
\usepackage{amssymb}
\usepackage{graphicx}
\usepackage{textcomp}
\usepackage{xcolor}
\usepackage{booktabs}
\usepackage{placeins}
\usepackage{enumitem}

\usepackage[ruled,vlined]{algorithm2e}
\include{pythonlisting}
\usepackage[pagebackref=true,breaklinks=true,letterpaper=true,colorlinks,bookmarks=false]{hyperref}

 \wacvfinalcopy 

\def\wacvPaperID{498} 
 
 \ifwacvfinal\fi
\begin{document}

\title{Answering Questions about Data Visualizations using Efficient Bimodal Fusion}

\author{Kushal~Kafle$^1$ \qquad Robik~Shrestha$^1$ \qquad Brian~Price$^{2}$ \qquad Scott~Cohen$^{2}$ \qquad Christopher~Kanan$^{1,3,4}$\\ 
$^1$Rochester Institute of Technology \qquad $^2$Adobe Research \qquad $^3$Paige \qquad $^4$Cornell Tech\\
{\tt\small $^1$\{kk6055, rss9369, kanan\}@rit.edu \qquad  $^2$\{bprice, scohen\}@adobe.com}
}

\newcommand{\commentout}[1]{}

\maketitle

\begin{abstract}

Chart question answering (CQA) is a newly proposed visual question answering (VQA) task where an algorithm must answer questions about data visualizations, e.g. bar charts, pie charts, and line graphs. CQA requires capabilities that natural-image VQA algorithms lack: fine-grained measurements, optical character recognition, and handling out-of-vocabulary words in both questions and answers. Without modifications, state-of-the-art VQA algorithms perform poorly on this task. Here, we propose a novel CQA algorithm called parallel recurrent fusion of image and language (PReFIL). PReFIL first learns bimodal embeddings by fusing question and image features and then intelligently aggregates these learned embeddings to answer the given question. Despite its simplicity, PReFIL greatly surpasses state-of-the art systems and human baselines on both the FigureQA and DVQA datasets. Additionally, we demonstrate that PReFIL can be used to reconstruct tables by asking a series of questions about a chart.
\end{abstract}

\section{Introduction}

Data visualizations such as bar charts, pie charts, and line graphs are common ways to present complex data in a manner that is easily interpretable to people. They are ubiquitous in both scientific and business documents. Data visualizations are designed to be effective at conveying trends and comparisons in a glance, while also preserving salient details. Using computer vision to parse these visualizations can enable extraction of information that cannot be gleaned by solely studying a document's text. Despite the high potential payoff and tremendous practical value, this problem has received little attention until recently. In 2018, two datasets for answering questions about data visualizations were introduced along with new algorithms~\cite{kafle2018dvqa,figureqa}; however, there is considerable room for improvement. Here, we propose a novel algorithm that exceeds the state-of-the-art on both of these datasets by a large margin.

\begin{figure}[t]
    \centering
    \includegraphics[width=0.47\textwidth]{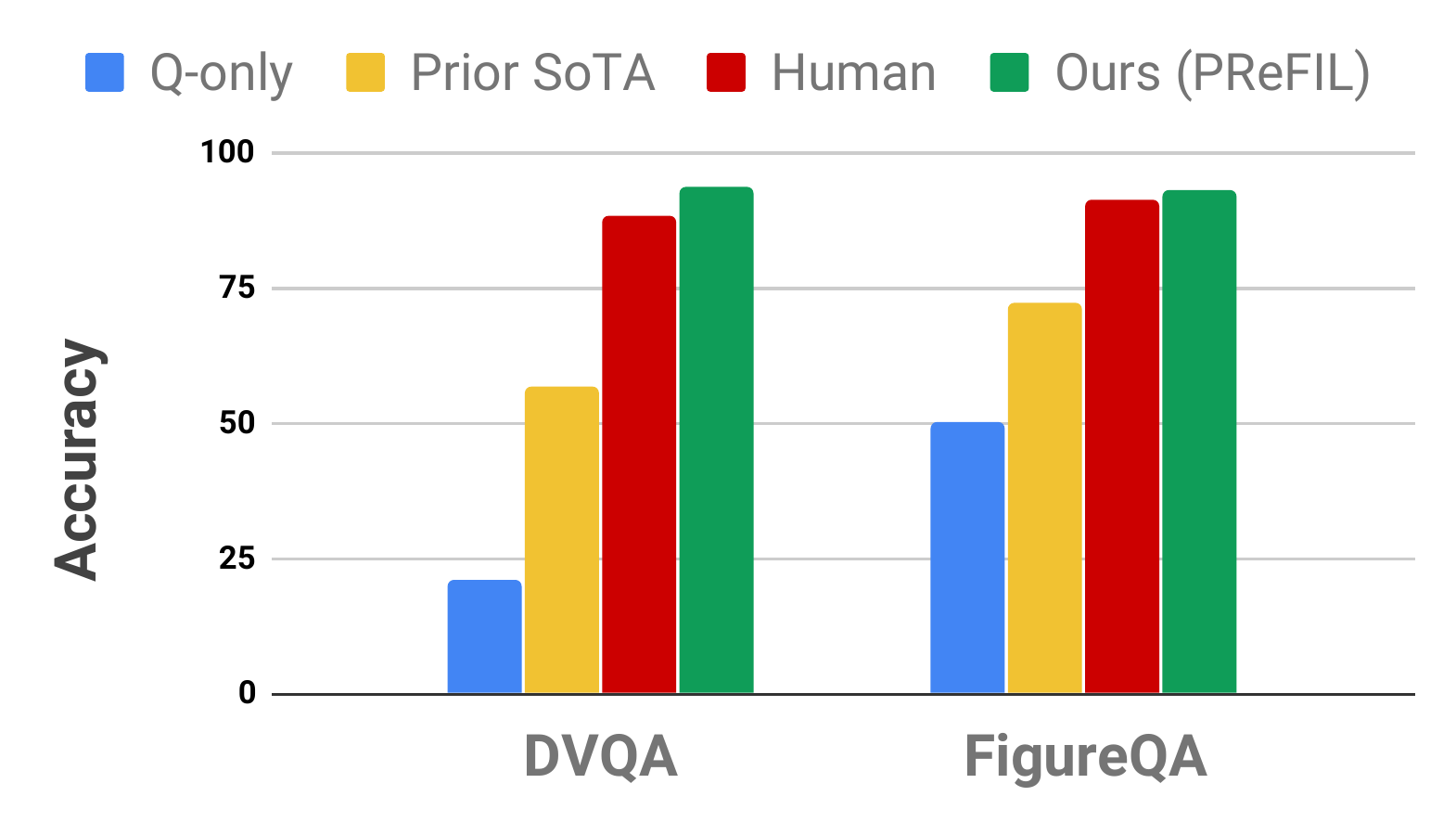}
    \caption{We propose the PReFIL algorithm for chart question answering (CQA). PReFIL surpasses the prior state-of-the-art (SoTA) and human baselines on DVQA and FigureQA datasets.}
    \label{fig:overview}
\end{figure}

Visual question answering (VQA) requires a system to answer questions about images~\cite{antol2015vqa,malinowski2014multi,kafle2016review,kafle2017data}. Several datasets for VQA has been proposed in recent years, include natural image understanding~\cite{malinowski2014multi,antol2015vqa}, counting~\cite{acharya2019tallyqa}, reasoning about synthetic scenes~\cite{johnson2016clevr}, medical image analysis~\cite{lau2018dataset}, scene text understanding~\cite{Singh2019TowardsVM}, and video comprehension~\cite{jang2017tgif}. Chart QA (CQA) is a VQA task involving answering questions about data visualizations. Formally, given an data visualization image $I$ and a question $Q$ about $I$, a CQA model must predict the answer $A$. CQA requires understanding of the relationships among  different `symbols' (elements in the chart) in an image. In contrast to natural images, even tiny modifications to the image can cause drastic changes in the correct answer, making CQA an excellent platform for studying reasoning mechanisms~\cite{figureqa,kafle2018dvqa}. CQA often requires optical character recognition (OCR) and handling words unique to a given visualization.

In this paper, we describe a novel algorithm called parallel recurrent fusion of image and language (PReFIL). PReFIL jointly learns bimodal embeddings by using both low- and high-level image features, which enable it to answer complex questions requiring multi-step reasoning and comparison without employing specialized relational or attention modules. Extensive experiments show that our algorithm outperforms current state-of-the-art methods, by a large margin in two challenging CQA datasets.

\vspace{1cm}

\begin{figure*}[t!]
    \centering
    \includegraphics[width=0.95\linewidth]{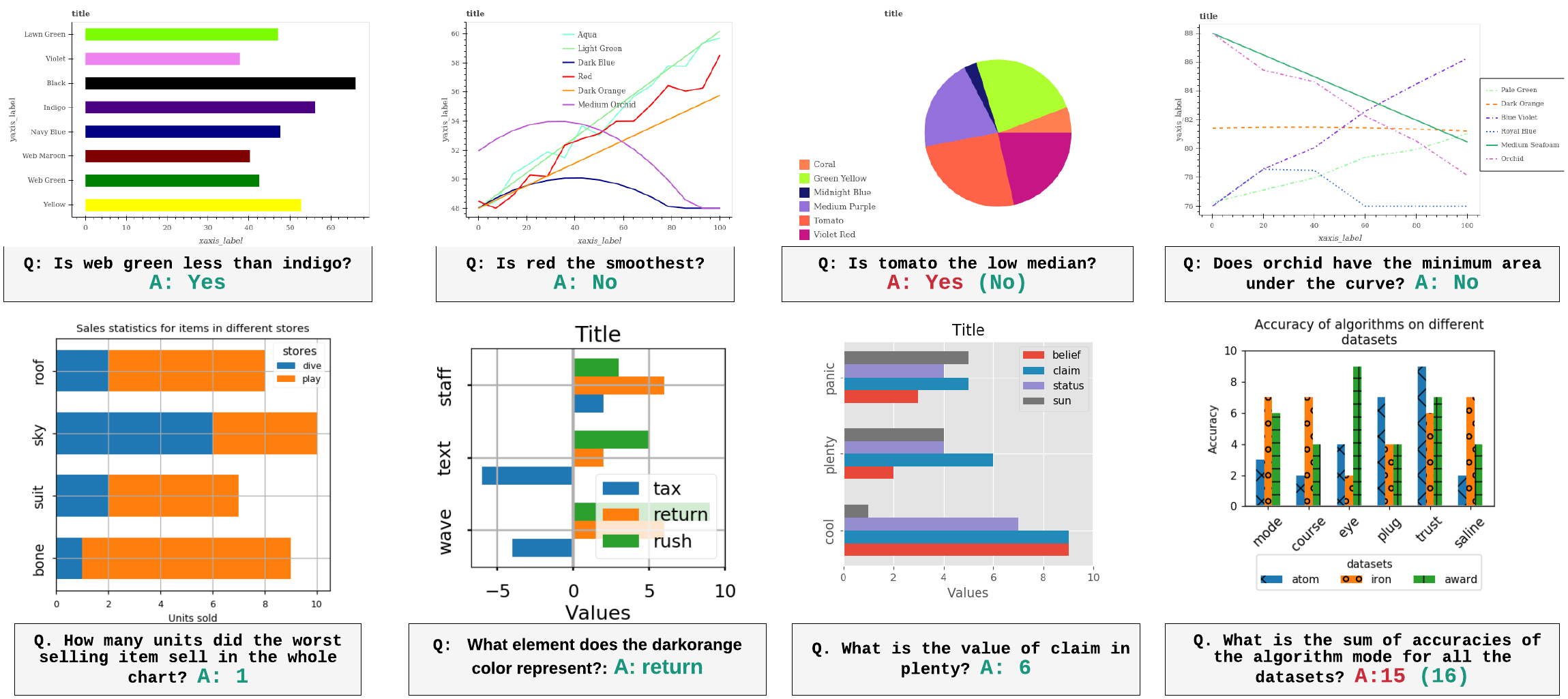}
    \caption{Example images and PReFIL outputs for FigureQA (top) and DVQA (bottom). Red denotes incorrect predictions. For incorrect predictions, correct answer is shown in parentheses. More examples are included in the supplementary materials. \label{fig:dvqa-examples}}
\end{figure*}

\paragraph{Our key contributions are:}
\begin{itemize}[noitemsep,nolistsep]
    \item We critically review existing CQA datasets outlining their strengths and weaknesses (Sec.~\ref{sec:datasets}).
    \item We collect human performance values for the DVQA dataset using crowd-sourcing (Sec.~\ref{sec:experiments}).
    \item We propose a novel algorithm called parallel recurrent early fusion of image and language (PReFIL) (Sec.~\ref{sec:PReFIL}). PReFIL greatly surpasses existing methods on CQA datasets and also outperforms humans on both DVQA and FigureQA (Sec.~\ref{sec:experiments}). PReFIL's code and pre-trained models will be publicly released.
    \item We pioneer the use of iterative question answering to reconstruct tables from charts (Sec.~\ref{sec:table-reconstruction}).
    \item In light of our results, we outline a road map toward creating more challenging datasets and algorithms for understanding data visualizations (Sec.~\ref{sec:discussion}).
\end{itemize}

\section{Related Work}

CQA is a form of VQA. Multiple natural image VQA datasets have been publicly released~\cite{malinowski2014multi,antol2015vqa,ren2015image,krishnavisualgenome,kafle2017analysis}. VQA has been explored in open-ended~\cite{antol2015vqa,kafle2016review}, counting~\cite{acharya2019tallyqa}, multiple choice~\cite{antol2015vqa,krishnavisualgenome}, and pointing type setups~\cite{visual7w,acharya2019vqd}. Most algorithms treat VQA as a classification problem in which the answer is a category~\cite{kafle2016review}. Several studies have shown that early natural image VQA datasets suffer from a high amount of bias, potentially making it easier for an algorithm to guess the answer without actually understanding of visual content~\cite{kafle2017analysis,AgrawalBP16,vqacp,kafle2019challenges}. As a remedy, some subsequent datasets have focused on synthetic scenes and diagrams where reasoning capacities can be better studied~\cite{AndreasRDK15,johnson2016clevr,kembhavi2016diagram,Kembhavi2017tqa}. 

 CQA requires capabilities not tested by other VQA tasks due to the innate differences in how information is presented in data visualizations~\cite{kafle2018dvqa,figureqa}. For instance, the information in charts is conveyed by only a small number of visual elements. Changes to even small image region (e.g., changing color of a legend entry) can drastically alter the information content of the whole chart whereas small changes in a natural image usually affects only a local region. This is one reason why algorithms designed for natural VQA have considerable difficulty when answering questions about data visualizations~\cite{kafle2018dvqa,figureqa}.

Another line of related work involves parsing of visual information in data visualization and other non-natural diagrams. There is a sizable body of prior work in this domain, ranging from extraction of visual elements in a chart~\cite{poco2017reverse,tsutsui2017data} to the extraction of underlying data~\cite{savva2011revision,kallimani2013extraction,cliche2017scatteract}. However, very limited work has been done in a question answering framework where multiple underlying abilities can be represented as a single task.

\subsection{Datasets for CQA}\label{sec:datasets}

\begin{table*}
\footnotesize
\centering
\caption{FigureQA vs. DVQA \label{tab:chartQA-datasets}}
\vspace{2pt}
\begin{tabular}{@{}cccccccc@{}}
\toprule 
6

         & Num. Images & Num. QA Pairs & Question Format & Chart Types & Number of Templates       & OCR & OOV  \\ \midrule
DVQA     & 300,000     & 3,487,194     & Open-ended      & 1            & 26 (Plus variations) & Required & Required     \\
FigureQA & 180,000     & 2,38,8698     & Yes/No          & 5            & 15 (No variations)    & Not Required    & Not Required \\ \bottomrule
\end{tabular}
\end{table*}

Two CQA datasets: DVQA~\cite{kafle2018dvqa} and FigureQA~\cite{figureqa}, are publicly available at the time of writing this paper. See Table~\ref{tab:chartQA-datasets} for their statistics. Example images are shown in Fig.~\ref{fig:dvqa-examples}. We briefly describe and compare both datasets.

\textbf{DVQA} has over 3 million question answer pairs for 300,000 images for bar charts. The question answer pairs in DVQA are divided into three categories: 1) structure understanding (e.g. ``How many bars are there?''), 2) data query (e.g., ``How many units of item $X$ were sold?''), and 3) reasoning (e.g. ``Is the accuracy of algorithm $X$ greater than algorithm $Y$?''). Since many questions refer to texts specific to the corresponding charts, systems must integrate OCR and dynamically expand their vocabulary to correctly answer questions. DVQA has two test splits: Test-Familiar and Test-Novel, with Test-Novel containing charts with texts that were not seen during training.

\textbf{FigureQA} has over 2 million question answer pairs for 180,000 images. It has five kinds of visualizations: 1) vertical bar charts, 2) horizontal bar charts, 3) pie charts, 4) line graphs and 5) dot-line graphs. Chart element colors are uniformly distributed in the training and validation sets. FigureQA has harder versions of the validation and test sets with color combinations that are unseen in the training set. Validation~1 and Test~1 have the same colors as the training set and Validation~2 and Test~2 have a color scheme that differs from training. Test set annotations are not publicly available. All questions are binary (yes/no) and demand multiple abilities, including finding the largest/smallest element (e.g. ``Is $X$ the largest/smallest?''), comparing values of two elements (e.g. ``Is $X$ greater/smaller than $Y$?''), and other scientific measurements (e.g. ``Does $X$ have maximum area under the curve?''). 

\subsubsection{DVQA versus FigureQA}
DVQA and FigureQA each have their own strengths and shortcomings. We compare and contrast them below.

\textbf{Shared strengths:} Both datasets are large and provide enough training samples to train large scale models, e.g. in DVQA, each unique visual element is repeated at least 1,000 times. Both datasets provide detailed annotations for all figure elements in addition to the question answer pairs, making it possible to create auxiliary tasks or use them as additional training signals. The creators of both datasets tried to eliminate some sources of bias. DVQA has randomized visual elements and it also has a balanced question answer distribution to make guessing difficult. Similarly, FigureQA has a randomized distribution of colors and a balanced distribution of ``yes'' and ``no'' answers for each unique question template. Lastly, both datasets provide both easy and hard test splits, where the hard test split measures generalization beyond what is seen during training. DVQA's ``Test Novel" split measures generalization to unseen words and FigureQA provides an ``alternated colors'' split where visual elements in the chart have different colors than the ones seen during training.


\textbf{DVQA's advantages:} In DVQA, questions about bars  are asked by referring to their text labels, e.g. ``What is the value of algorithm $X$?'' where $X$ is an actual label in the chart and it will be different for each chart even if they have the same appearance, e.g. identical red bars may have label $X$ in one image and $Y$ in another. This requires integrating OCR into the system. In contrast, FigureQA refers to chart elements by their color, e.g. ``red bars'' will always be referred to as ``red'' making it easier for systems to identify a chart's elements. Since DVQA uses chart labels, algorithms must take into account that some of the words may be out-of-vocabulary (OOV) and unseen during training for both questions and answer. To handle this, systems need to have a vocabulary that can be dynamically adjusted during testing. FigureQA has no OOV answers. DVQA also tests for more tasks than FigureQA. For bar charts, DVQA contain most of the tasks in FigureQA (e.g. identifying colors, comparing values, etc.) and several that are not required for FigureQA (e.g. data measurement and inferring structure of the chart). Finally, while DVQA contains only bar charts, its bar charts have increased visual complexity compared to those in FigureQA. FigureQA is limited to single-variable vertical and horizontal bar charts, whereas DVQA also has grouped bar charts and stacked bar charts with legends. DVQA's bars can be hatched, monochrome, and have negative values, all of which are absent in FigureQA.

\textbf{FigureQA's advantages:} While DVQA has only bar charts, FigureQA has three kinds of data visualizations: bar charts, pie charts, and line graphs. This allows FigureQA to have unique question-types that are not encountered for bar chart alone. E.g., for line graphs, FigureQA requires determining the area under the curve, and whether one line intersects another. These are not tested in DVQA. FigureQA also tests compositional reasoning by asking questions about unknown color combinations in chart elements, whereas colors are randomly distributed in DVQA.

\textbf{Shared limitations:} As synthetically generated datasets, both DVQA and FigureQA omit much of the variability found in real-world data visualizations. All of DVQA's charts were made with Matplotlib  and all of FigureQA's were made with Bokeh. The variation introduced is limited to the capabilities of these packages. FigureQA uses only generic titles and other chart elements. DVQA has some variety but ultimately is limited to a few templates. Likewise, both datasets have formulaic, templated questions. While questions can be complex, they lack the diversity of human generated queries. In the discussion we elaborate further on how future datasets could overcome these limitations. 

\subsection{Existing CQA Algorithms}\label{sec:algorithms}

For DVQA, in \cite{kafle2018dvqa} SANDY (SAN with DYnamic encoding) model was proposed. SANDY used a modified version of the stacked attention network (SAN)~\cite{Yang2016,kazemi2017show}, which has been widely used for VQA~\cite{kazemi2017show,anderson2018bottom}. SAN uses the question to apply attention to the convolutional feature maps. It cannot handle DVQA's OOV words in its test set or the chart specific words found in its questions and answers. To address this, SANDY uses an off-the-shelf OCR method to recognize such words and introduced dynamic encoding to represent OOV and chart-specific words. SANDY's dynamic encoding scheme for OCR can be incorporated into any classification-based VQA algorithm.

FigureQA's creators used a relation network (RN)~\cite{santoro2017simple} on their dataset.  RN encodes pairwise interactions between every pair of ``objects'' in an image, enabling it to answer questions involving relationships. Each ``object'' is a cell of a convolutional feature map. RN has been shown to be especially effective at compositional reasoning in CLEVR~\cite{santoro2017simple}, and it exceeded baselines on FigureQA.

FigureNet~\cite{reddy2018question} is a multi-step algorithm for FigureQA composed of different modules. The first module is called the spectral segregator, which identifies the elements and colors of the chart. It is followed by the extraction module, which  quantifies the values represented by each element. This is then used with a feed-forward network to predict the answer. FigureNet uses the detailed annotations of FigureQA's chart elements to pre-train each of the modules. Because FigureNet relies on having access to the measurements of each chart element, they could only apply it to FigureNet's bar and pie charts.

To assess bias in their datasets, the creators of  FigureQA and DVQA both studied question-blind and image-blind models. They found that these models performed abysmally indicating that vision and language must be jointly used to correctly answer the questions. The creators of both datasets also tested simple question+image fusion schemes. These  worked better than the blind baselines, but this did not suffice for handling the complexity found in CQA. This is in contrast to VQA with natural images, where these algorithms fare comparatively well.

Compared to existing work, our model does not employ complex attention or relational modules, and unlike FigureNet, it does not require additional supervised annotations for training on FigureQA.

\begin{figure*}
    \centering
    \includegraphics[width=0.86\textwidth]{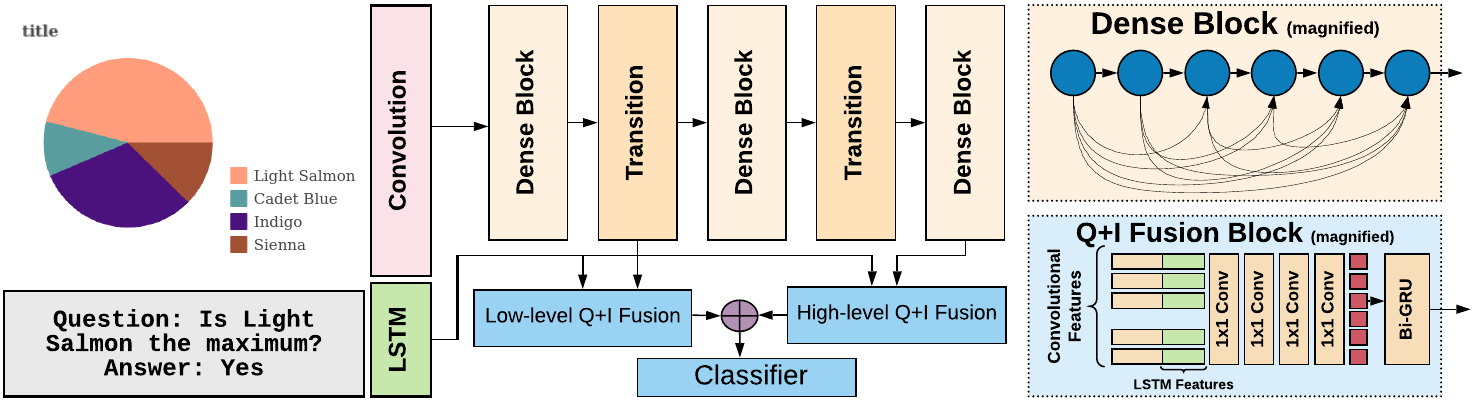}
    \caption{Components of our PReFIL model. Magnified views show the details of each dense block and Q+I fusion block. \label{fig:our-model}}
\end{figure*}

\section{The PReFIL Model}\label{sec:PReFIL}

We propose the PReFIL algorithm for CQA. As shown in Fig.~\ref{fig:our-model}, PReFIL has two parallel Q+I fusion branches. Each branch takes in question features (from an LSTM) and image features from two locations of a 40-layer DenseNet, \ie low-level features (from layer 14) and high-level features (from layer 40). Each Q+I fusion block concatenates the question features to each element of the convolutional feature map, and then it has a series of $1 \times 1$ convolutions to create question-specific bimodal embeddings. These embeddings are recurrently aggregated and then fed to a classifier that predicts the answer. Despite being composed of relatively simple elements, PReFIL outperforms more complex methods that use RNs and attention mechanisms. The three main stages of PReFIL are  described in the next subsections. For DVQA, an additional fourth OCR-integration component is required (Sec. \ref{subsec:OCR}). In Sec.~\ref{sec:ablation}, we conduct studies to understand the value of each stage.

\subsection{Multi-stage Image Encoder}
For all model variants, image encoder is a DenseNet~\cite{huang2017densely} trained from scratch. DenseNet is an efficient architecture for training deep convolutional neural networks (CNNs). It is comprised of several ``dense blocks'' and ``transition blocks'' between the dense blocks. Each dense block has several convolutional layers, where each layer uses outputs of all preceding layers as its input. The transition block sits between two dense blocks and serves to change feature-map sizes via convolution and pooling. This architecture encourages feature reuse, improves training, and mitigates vanishing-gradients, making it easy to train very deep networks. Feature reuse allows DenseNet to learn complex visual features with fewer parameters compared to other architectures~\cite{huang2017condensenet}.

In deep CNNs, complex features are learned as a hierarchy of visual features with earlier layers learning simple features and later layers learning higher-level features that are combinations of simpler features~\cite{yosinski2015understanding}. In data visualizations, simpler features such as color patches, lines, textures, etc. convey important information that is often abstracted away by deeper layers of a CNN. Hence, we use both low- and high-level convolutional features in our model, both of which are fed to parallel fusion module alongside question embeddings learned using an LSTM. We study the importance of both low and high level features in Sec.~\ref{sec:ablation}.

\subsection{Parallel Fusion of Image and Language} Jointly modulating visual features using  vision and language features can allow models to learn richer features for downstream tasks~\cite{malinowski2018visual,perez2018film,shrestha2019ramen}. Our Q+I fusion block does this by first concatenating all of the input convolutional feature map's spatial locations with the question features, and then bimodal fusion occurs using a series of layers that use $1 \times 1$  convolutions~\cite{malinowski2018visual,shrestha2019ramen}. This allows the question to modulate visual feature processing and yields bimodal embeddings that capture information from both the image and the question. This approach resembles early VQA models that concatenated CNN embeddings to question embeddings, with the critical difference being that this happens before spatial pooling across the entire scene. We do this for both low-level and high-level convolutional features in parallel. In Sec.~\ref{sec:ablation}, we study the importance of learning bimodal embeddings jointly. 

\subsection{Recurrent Aggregation of bi-modal features} 
 In CNNs, the most common approach to aggregating information from a feature map $F \in \mathbb{R}^{M\times N \times D}$ is to collapse across the spatial dimensions to produce a $D$ dimensional vector by mean pooling or max pooling. An alternative is to ``flatten'' $F$ to turn it into a $DMN$-dimensional vector. Recent attentive approaches have  explored using a weighted sum, where the relative importance of each region is based on the question. These methods may fail to  capture \textit{interactions} among features, especially for high-level tasks such as question answering. To address this, we aggregate information using a bidirectional gated recurrent unit (bi-GRU), which sequentially takes in the $D$-dimensional features from each of the $MN$ locations in $F$. The aggregated features are sent to a classifier to predict the answer. As ablation, we also try sum-pooling for aggregation in Sec.~\ref{sec:ablation}.

\subsection{OCR Integration for DVQA dataset}\label{subsec:OCR}

Unlike FigureQA and most VQA tasks, DVQA requires OCR to answer its reasoning and data questions.  A fixed vocabulary consisting of all the words seen during training is not enough since the model will encounter OOV words during testing. To integrate OCR into PReFIL, we use the same dynamic encoding scheme used by the SANDY model~\cite{kafle2018dvqa}. Dynamic encoding creates an image specific dictionary that associates the spatial positions of scene elements with entries in the dictionary.  Before running the net, all words are detected using OCR and then they are associated with the appropriate element in the dynamic encoding dictionary based on each word's spatial position. Subsequently, if a question word is encountered that is in the dynamic dictionary then the appropriate element is set to 1. For answers, a portion of the classification layer is reserved for the dynamic encoding outputs. See \cite{kafle2018dvqa} for additional details.

To assess impact of OCR, we test three OCR versions as well as a version of algorithm trained without the dynamic encoding, i.e., only using a fixed-vocabulary constructed from the train split. The first two OCR systems are identical to those used by \cite{kafle2018dvqa}: an oracle (perfect) OCR model and a real OCR system using Tesseract. Because Tesseract has been found to be sub-optimal when used directly on diagrams~\cite{kembhavi2016diagram}, we also study using a two-stage OCR pipeline where we first detect text and then run OCR on the detected regions to recognize the text. Specifically, we use the EAST text detector~\cite{zhou2017east} to detect text-regions for images rotated at 0, 45 and 90 degrees. We then perform non-maximum suppression on overlapping detections and crop them. Each cropped region is resized by 200\% and sent to the Tesseract OCR to obtain the text within each region. The rest of the dynamic encoding scheme remains unchanged.

\subsection{Model and Training Hyperparameters}

\textbf{Question Encoding:} Question words are represented by 32 dimensional learned word embedding and passed through an LSTM which provides a 256-dimensional embedding representing the whole question.

\textbf{DenseNet:} We use a 40 layer DenseNet composed of 3 dense blocks with 12 layers each. The number of initial filters is 64 and the growth rate is set to 32.

\textbf{Preprocessing:} DVQA images are resized to a size of $256\times 256$. FigureQA images are all differently sized but we resize them to $320\times 224$ which maintains an \textit{average} width-height aspect ratio. For data augmentation during training, both DVQA and FigureQA images are padded with 8 pixels  on all sides, followed by random crops and random rotations of up to 3 degrees.

\textbf{Q+I Fusion:} Inputs to Q+I block are batchnormed. Each Q+I fusion block is composed of four $1\times1$ convolutions with 256 channels and ReLU. 

\textbf{Recurrent Fusion:} The bimodal features are aggregated using a 256 dimensional bi-directional GRU. The forward and backward direction outputs are concatenated to form a 512 dimensional vector which is fed to the classifier.

\textbf{Classifier:} The aggregated bimodal features are projected to a 1024 fully connected ReLU layer, which was regularized using  dropout of 0.5 during training. The classification layer is binary for FigureQA. For DVQA, the classification layer has 107 units,  with 77 units for predicting `common' answers such as `yes', `no', `three groups', etc, and 30 special tokens for predicting answers that require OCR, which allows PReFIL to produce OOV answer tokens that are unseen during training (see Sec.~\ref{subsec:OCR} for details).

\textbf{Losses and Optimizers:} For DVQA, PReFIL is trained using multinomial cross-entropy loss. For FigureQA, PReFIL is trained using binary cross entropy loss. Following~\cite{kim2018bilinear}, we use Adamax optimizer with a gradual learning rate warm-up, with a base learning rate of $7\times 10^{-4}$. The first 4 epochs use a learning rate of $(0.5\times epoch\times base)$ and the rate starts decaying by a factor of 0.7 from epochs 15 to 25. For DVQA, all models are trained for a fixed 25 epochs. For FigureQA, we train them until they converge on the validation set and submit predictions to its creators for assessment on the non-public test set.

\begin{table*}
\centering
\footnotesize
\label{tab:results-figureqa}
\caption{Results for the FigureQA dataset for our PReFIL algorithm compared to baseline and existing algorithms.}
\vspace{2pt}
\begin{tabular}{@{}lcccccccccccc@{}}
\toprule
          & \multicolumn{6}{c}{\textbf{Validation 1 - Same Colors}}         & \multicolumn{6}{c}{\textbf{Validation 2 - Alternated Colors}} \\ 
          \cmidrule(r){2-7} \cmidrule(r){8-13}
          & vBar  & hBar  & Pie   & Line & Dot-line & Overall        & vBar  & hBar  & Pie   & Line  & Dot-line & Overall        \\ \midrule
QUES \cite{figureqa}      & -     & -     & -     & -    & -        & -              & -     & -     & -     & -    & -                 & 50.01          \\
IMG+QUES \cite{figureqa}  & 61.98     & 62.44     & 59.63     & 57.07    & 57.35        & 59.41              & 58.60 & 58.05 & 55.97 & 56.37 & 56.97    & 57.14          \\
RN \cite{figureqa}        & 85.71     & 80.60     & 82.56     & 69.53    & 68.51        & 76.39              & 77.35 & 77.00 & 74.16 & 67.90 & 69.04    & 72.54          \\
FigureNet \cite{reddy2018question} & 87.36 & 81.57 & 83.13 & -    & -        & -          & -     & -     & -     & -     & -        & -              \\
PReFIL (Ours)      & \textbf{98.80}   & \textbf{98.09}   & \textbf{95.11}   & \textbf{91.82}  & \textbf{92.19}      & \textbf{94.84} & \textbf{98.46}   & \textbf{97.94}   & \textbf{93.57}   &88.50   &\textbf{ 90.30}      & \textbf{93.26} \\ \midrule
& \multicolumn{6}{c}{\textbf{Test 1 - Same Colors}}         & \multicolumn{6}{c}{\textbf{Test 2 - Alternated Colors}}   \\ \midrule
PReFIL (Ours)      & 98.79   & 98.14   & 95.35   & 91.98  & 92.05      & 94.88 & 98.41   & 97.93   & 93.58   & 88.26  & 90.07      & 93.16 \\ \bottomrule
\end{tabular}
\end{table*}

\section{Experiments and Results}
\label{sec:experiments}

\begin{table}[]
\footnotesize
\centering
\caption{Results on FigureQA's Test 2 split with alternated color schemes. All results are from the 16,876 questions answered by human annotators.}
\vspace{2pt}
\label{tab:results-figureqa-test}
\begin{tabular}{@{}lcccc@{}}
\toprule
Type & \multicolumn{1}{c}{PReFIL(Ours)} & \multicolumn{1}{c}{Q+I \cite{figureqa}} & \multicolumn{1}{c}{RN \cite{figureqa}} & \multicolumn{1}{c}{Human \cite{figureqa}} \\ \midrule
vBar & \textbf{98.25} & 59.63 & 77.13 & 95.90 \\
hBar & \textbf{97.98} & 57.69 & 77.02 & 96.03 \\
Pie & \textbf{92.84} & 55.32 & 73.26 & 88.26 \\
Line & 87.79 & 54.46 & 66.69 & \textbf{90.55} \\
Dot-line & \textbf{89.57} & 54.19 & 69.22 & 87.20 \\ \midrule
Overall & \textbf{92.79} & 56.04  & 72.18 & 91.21 \\ \bottomrule
\end{tabular}
\end{table}







\begin{table*}
\centering
\footnotesize
\caption{Results for the DVQA dataset for PReFIL compared to baselines and existing algorithms.}
\vspace{3pt}
\label{tab:results-dvqa}
\begin{tabular}{@{}lcccccccc@{}}
\toprule
            & \multicolumn{4}{c}{\textbf{Test-Familiar}} & \multicolumn{4}{c}{\textbf{Test-Novel}} \\ 
            \cmidrule(r){2-5} \cmidrule(l){6-9}
            & Structure  & Data   & Reasoning  & Overall & Structure  & Data  & Reasoning & Overall \\ \midrule
QUES \cite{kafle2018dvqa}        & 44.03      & 9.82   & 25.87      & 21.06   & 43.90      & 9.80  & 25.76     & 21.00   \\
IMG+QUES \cite{kafle2018dvqa}    & 90.38      & 15.74  & 31.95      & 32.01   & 90.06      & 15.85 & 31.84     & 32.01   \\ 

\midrule
SANDY (No OCR) \cite{kafle2018dvqa}  & 94.71      & 18.78  & 37.29      & 36.02   & 94.82 &18.92& 37.25& 36.14 \\
PReFIL (No OCR) & 99.77           & 23.39       &  49.05   & 47.70 &    99.77     &     23.43       &  49.21     & 47.86   \\

\midrule
SANDY (Tesseract OCR) \cite{kafle2018dvqa}  & 96.47      & 37.82  & 41.50      & 45.77   & 96.42      & 37.78 & 41.49     & 45.81  \\ 
PReFIL (Ours, Tesseract OCR) & 99.75           & 49.00       &  74.61   & 69.63 &    99.73     &     48.91       &  74.07     & 69.53         \\
PReFIL (Ours, Improved OCR) & 99.73           & 68.55       &  83.44 & 80.88 &    99.57     &     67.13       &  80.73     & 80.04         \\

\midrule
SANDY (Oracle OCR) \cite{kafle2018dvqa}  & 96.47      & 65.40  & 44.03      & 56.48   & 96.42      & 65.55 & 44.09     & 56.62   \\
PReFIL (Ours, Oracle OCR) & \textbf{99.77}      & \textbf{95.80}  & \textbf{95.86}    & \textbf{96.37}   & \textbf{99.78}      & \textbf{96.07} & \textbf{95.99}     & \textbf{96.53}   \\ \midrule
Human  & -      & -  & -      & -   & 96.19      & 88.70 & 85.83     & 88.18   \\ \bottomrule
\end{tabular}

\end{table*}

\subsection{FigureQA}

FigureQA has two validation sets and two non-publicly available test sets. Validation~1 and Test~1 have the same colors as the training set and Validation~2 and Test~2 have a color scheme that differs from training. Test sets are not publicly available and the results were obtained by sending the predictions to the authors. Existing works do not report accuracy for the full test set, but we report results for both validation and test sets in Table~\ref{tab:results-figureqa} for completeness.

Our PReFIL algorithm exceeds FigureNet by a large margin despite FigureNet having access to additional annotations. FigureNet is incapable of answering questions about line and dot-line graphs, so it is only evaluated on vBar, hBar and Pie. For these chart types, average accuracy for FigureNet is 83.9\%, compared to 97.33\% for ours.

 FigureQA also provides human performance for a \textit{subset} of Test~2, which is not available for the other sets. We report PReFIL's performance compared to other baselines and human performance on the exact same subset in Table~\ref{tab:results-figureqa-test}. PReFIL outperforms the human baseline for four out of five categories and also surpasses overall human accuracy. When analyzed for different question templates, PReFIL outperforms humans  for 12 out of 15 question templates. PReFIL shows the most improvements for questions requiring  measurements, e.g. for the question template  ``Is X the high/low median?'' PReFIL outperforms human accuracy by over 7\% (absolute). Detailed results for all 15 templates are presented in the supplementary materials.

\subsection{DVQA}

DVQA is split into Test-Familiar, which contains bar charts with words that are also encountered in its Train set, and Test-Novel, which contains bar charts with novel words in them. Results for both DVQA splits are given in Table~\ref{tab:results-dvqa}. PReFIL surpasses SANDY by over 40\% in accuracy when both the baseline SANDY and our PReFIL method have access to a perfect Oracle OCR, which is emulated by providing the correct text-annotations for all the elements in the images. When using Tesseract OCR, we obtain about a 24\% improvement overall on both test sets. To demonstrate that PReFIL's performance scales with access to better OCR, we also test a version that uses an improved OCR pipeline (see Sec.~\ref{sec:PReFIL}). This further improves PReFIL's performance by about 11\% bringing it closer to the results of the oracle OCR version. When OCR is removed entirely, PReFIL still performs about 11\% better than SANDY without OCR, but this ablation renders many data and reasoning questions impossible to answer.  This re-affirms the assertion by DVQA's creators that OCR integration is essential for answering the data and reasoning questions in the dataset~\cite{kafle2018dvqa}. 

Across all OCR variants, PReFIL outperforms SANDY. Moreover, PReFIL's performance scales much better when better OCR is available: 11\% gain for SANDY vs. 26\% gain for PReFIL when moving from the imperfect Tesseract OCR setup to the perfect Oracle OCR setup. Our results show that PReFIL is as effective for novel words (Test-Novel) as it is for familiar words (Test-Familiar). This is enabled by the dynamic OCR integration, which is designed to be agnostic to whether a word has been encountered before.

Because no human accuracy estimate for DVQA existed, we had people answer 5000 randomly selected questions for 5000 images from the DVQA Test-Novel split. The annotators were shown example QA pairs from each of three question types. We perform post processing on the provided answers to rectify minor answer entry errors. First, we found some annotators used decimal points or spelled out numerals (``5.0'' or ``five'' instead of ``5'') despite our instructions to only use integers when answers are numbers. Because DVQA contains only integers, we convert all such occurrences to the nearest integer. For word answers, we allow one character typographic error to be discounted. Results for humans and models are given in Table~\ref{tab:results-dvqa}. With perfect OCR, PReFIL surpasses the DVQA human accuracy result across question types. Its performance on reasoning questions is  almost 10\%  greater (absolute), and it exceeds them by almost 8\% (absolute) for DVQA's data questions, which require measurement. However, without perfect OCR humans exceed PReFIL, although the better OCR used for PReFIL does lead to significantly better results than PReFIL with improved OCR. This suggests that the underlying core algorithm and reasoning mechanisms in PReFIL work well for DVQA, and the main limiting factor is OCR.

\subsection{Ablation Studies}
\label{sec:ablation}
We studied the contribution of PReFIL's components by analyzing  a series of ablation models. We trained each model variation and the original PReFIL (Oracle OCR) for 25 epochs on a subset of DVQA that has only 500,000 randomly selected training samples. The ablation models  are:
\begin{itemize}[noitemsep,nolistsep]
    \item \textbf{No bimodal embeddings:} Instead of learning bimodal embeddings, the question is concatenated after the recurrent aggregation and fed to the classifier.
    \item \textbf{No low-level features:} Only the high-level (layer 40 output) DenseNet features are used.
    \item \textbf{No high-level features:} Only the low-level (layer 14 output) DenseNet features are used. This is equivalent to using a shallower DenseNet.
    \item \textbf{No recurrent aggregation: } Instead of recurrent aggregation, output is aggregated via summation.
\end{itemize}

As shown in Table~\ref{tab:ablation}, all of PReFIL's components impact its performance. Removing bimodal embeddings causes the largest accuracy drop (over 12\% absolute). The next largest is caused by removing low and high-level visual features (1.3\% and 6\% absolute). 

\begin{table}
\centering
\footnotesize
\caption{PReFIL ablation studies on a 500K DVQA train subset. \label{tab:ablation}}
\vspace{3pt}
\begin{tabular}{lcc}
\toprule
Ablation Model         & Test Familiar & Test Novel \\ \midrule
PReFIL (full model) & \textbf{91.18}          & \textbf{91.32}           \\
No bimodal embedding       &      78.00         &  78.36         \\
No high-level features & 85.68              & 85.86            \\
No low-level features & 89.87              & 90.05            \\
No recurrent aggregation   & 90.88             &    91.14        \\ \midrule
\end{tabular}
\end{table}

\subsection{Table Reconstruction by Asking Questions}
\label{sec:table-reconstruction}

\commentout{
\begin{algorithm}[t!]
\footnotesize
 \eIf{bar\_type is single}{
    n = ans("How many bars are there?")\;
    \For{$i\gets1$ to $n$}{ 
        data[i] = ans(``What is the value of the $i^{th}$ bar?")\;
        label[i] = ans(``What is the label of the $i^{th}$ bar?")\;
    }
  }{
    m = ans(``How many groups are there?")\;
    n = ans(``How many bars are there per group?")\;
    \For{$i\gets1$ to $m$}{ 
        bar\_label[i] = ans(``What is the label of the $i^{th}$ group?")\;
        \For{$j\gets1$ to $n$}{ 
            \If{i==1}{
                legend\_label[j] = ans(``What is the label of the $j^{th}$ bar in each group?")\;
            }
            data[i,j] = ans(``What is the value of the $j^{th}$ bar in $i^{th}$ group?")\;
        }
    }
}
\caption{Iterative QA for Data Reconstruction}
\label{alg:reconstruction}
\end{algorithm}
}

\begin{algorithm}[t!]
 \eIf{bar\_type is single}{
    n = ans("How many bars are there?")\;
    \For{$i\gets1$ to $n$}{ 
        data[i] = ans(``What is the value of the $i^{th}$ bar?")\;
        label[i] = ans(``What is the label of the $i^{th}$ bar?")\;
    }
  }{
    m = ans(``How many groups are there?")\;
    n = ans(``How many bars are there per group?")\;
    \For{$j\gets1$ to $n$}{ 
        legend\_label[j] = ans(``What is the label of the $j^{th}$ bar in each group?")\;
    }    
    \For{$i\gets1$ to $m$}{ 
        bar\_label[i] = ans(``What is the label of the $i^{th}$ group?")\;
        \For{$j\gets1$ to $n$}{ 
            data[i,j] = ans(``What is the value of the $j^{th}$ bar in $i^{th}$ group?")\;
        }
    }
}
\caption{Iterative QA for Data Reconstruction}
\label{alg:reconstruction}
\end{algorithm}

\begin{table}
\centering
\footnotesize
\caption{Bar chart reconstruction accuracy (\%) using  Algorithm~\ref{alg:reconstruction} with PreFIL (Oracle OCR).}
\vspace{3pt}
\label{tab:table-reconstruction}
\begin{tabular}{@{}lcc@{}}
\toprule
               & Test Familiar & Test Novel \\ \midrule
 Shape Prediction    &    99.97       & 99.97            \\
 Label Prediction    &   97.78            &   97.78          \\
 Value Prediction  &     84.21          &   84.75         \\ \midrule
 Overall  &     90.79          &   91.10         \\ \midrule

 \end{tabular}
\end{table}

\begin{figure}[t!]
    \centering
    \includegraphics[width=0.99\linewidth]{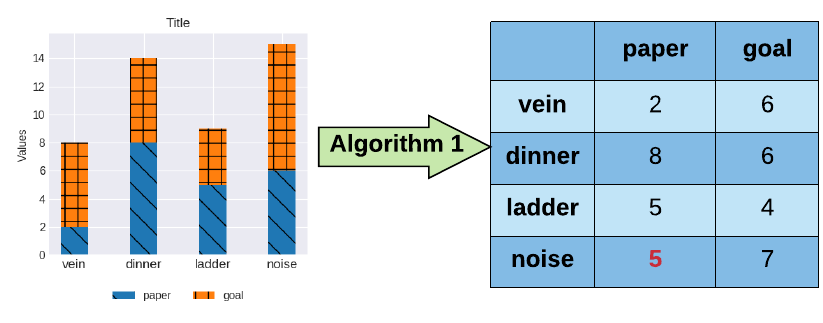}
    \caption{An example output of the chart to table algorithm. Red denotes incorrect predictions.}
    \label{fig:table-reconstruction-example}
\end{figure}

We introduce table reconstruction for DVQA as an application of PReFIL.  DVQA's question templates provide the questions needed to completely reconstruct its bar charts by iteratively asking questions about each chart. Our approach is given in Algorithm~\ref{alg:reconstruction}. An example reconstruction is shown in Fig.~\ref{fig:table-reconstruction-example}, and results using PReFIL (Oracle OCR) are given in Table~\ref{tab:table-reconstruction}. Shape prediction can be done with near perfect accuracy, but there is a drop in performance for both label and value prediction.  To study the accuracy of different components in chart reconstruction, we also report accuracy on three main components of the iterative question-answering: 1) Shape prediction: Questions about number of bars and legends in the picture; 2) Label prediction: Predicting the label of given bar or legend; and 3) Value Prediction: Predicting the value of a given bar.

\section{Discussion}
\label{sec:discussion}

PReFIL surpassed prior state-of-the-art methods for both DVQA and FigureQA. While PReFIL exceeded the human baseline for FigureQA, results are more nuanced for DVQA due to OCR model variations. All OCR versions  exceeded the human baseline for structure questions, but only PReFIL using oracle OCR exceeded humans across all question types. We found  that better OCR methods led to better results for DVQA. Future developments in OCR technology would likely improve PReFIL further. 

The strong results in this paper suggest that the community is ready for more difficult CQA datasets. We have the following recommendations:
\begin{itemize}[noitemsep,nolistsep]
    \item \textbf{Charts in the wild:} The charts in FigureQA and DVQA were methodologically generated, but human-generated charts in real-world business and scientific documents can contain variations that these datasets omit. Additional text in the chart or human annotations would likely cause the dynamic encoding method used by PReFIL to fail. Next generation datasets should contain charts extracted from real-world documents.

    \item \textbf{Human generated questions:} The questions in both FigureQA and DVQA were created with templates, which do not capture all the nuances of natural language. Deploying a chart question answering system will require it to handle human-generated queries. Studies on the synthetically generated CLEVR dataset have demonstrated that algorithms experience a large drop in performance when natural language questions are asked to a model trained only on CLEVR~\cite{clevr-iep}.  Future CQA datasets should include human-generated question-answer pairs. 
    
    \item \textbf{Document-level CQA:} FigureQA and DVQA have well-defined image regions and all information needed to answer a question is contained in that image. To understand charts in documents, information in the rest of the document may be necessary to answer questions about the chart. Beyond typical CQA algorithm abilities, this requires document question answering~\cite{Clark2018SimpleAE}, page segmentation~\cite{he2017multi}, and more. Creation of such a dataset would greatly increase the challenge for future algorithms and better match real-world usage. 
\end{itemize}

\section{Conclusion}
We proposed PReFIL, a new CQA system that improves the state-of-the-art and surpasses human accuracy on two datasets. Like other VQA tasks~\cite{kafle2019challenges}, our results suggest  harder datasets are needed. For CQA, better OCR is also important for advancing the field. Our work has the potential to improve retrieval of information from charts, which has numerous applications, including automatic information retrieval, table reconstruction, and enabling better understanding of charts by people with visual impairments.
\vspace{-3px}
\paragraph{Acknowledgements.} This research was supported in part by NSF award \#1909696 to C.K. We thank NVIDIA for gifting a GPU to C.K.'s lab. This work was supported in part by a gift from Adobe Research.

{\small
\bibliographystyle{ieee}
\bibliography{egbib}
}

\appendix

\begin{table*}
\centering
  \caption{Results for PReFIL compared with RN~\cite{santoro2017simple,figureqa} and Human baseline~\cite{figureqa} compared with each unique question template in FigureQA.}
  \vspace{1cm}
  \label{tab:extra-results}
\begin{tabular}{@{}llccc@{}}
\toprule
\textbf{Question Template}                    & \textbf{Figure Types} & \textbf{RN~\cite{santoro2017simple,figureqa}}               & \textbf{Human~\cite{figureqa}}            & \textbf{PReFIL (Ours)}                      \\ \midrule
Is X the minimum?                             & bar, pie              & 76.78                     & 97.06                     & \textbf{97.20}                     \\
Is X the maximum?                             & bar, pie              & 83.47                     & 97.18                     & \textbf{98.07}                     \\
Is X the low median?                          & bar, pie              & 66.69                     & 86.39                     & \textbf{93.07}                     \\
Is X the high median?                         & bar, pie              & 66.50                     & 86.91                     & \textbf{93.00}                     \\
Is X less than Y ?                            & bar, pie              & 80.49                     & 96.15                     & \textbf{98.20}                     \\
Is X greater than Y ?                         & bar, pie              & 81.00                     & 96.15                     & \textbf{98.07}                     \\
Does X have the minimum area under the curve? & line                  & 69.57                     & \textbf{94.22}            & 94.00                              \\
Does X have the maximum area under the curve? & line                  & 78.45                     & 95.36                     & \textbf{96.91}                     \\
Is X the smoothest?                           & line                  & 58.57                     & \textbf{78.02}            & 71.87                              \\
Is X the roughest?                            & line                  & 56.28                     & \textbf{79.52}            & 74.67                              \\
Does X have the lowest value?                 & line                  & 69.65                     & 90.33                     & \textbf{92.17}                     \\
Does X have the highest value?                & line                  & 76.23                     & 93.11                     & \textbf{94.83}                     \\
Is X less than Y?                             & line                  & 67.75                     & 90.12                     & \textbf{92.38}                     \\
Is X greater than Y?                          & line                  & 67.12                     & 89.88                     & \textbf{92.00}                     \\
Does X intersect Y ?                          & line                  & 68.75                     & 89.62                     & \textbf{91.25}                     \\ \midrule
Overall                                       & bar,pie,line                     & 72.18 & 91.21 & \textbf{92.79} \\ \bottomrule
\end{tabular}
\label{fig:frontpage}
\end{table*}

\section{Analysis per FigureQA Question Template}

Table \ref{tab:extra-results} shows results for PReFIL compared to RN~\cite{santoro2017simple,figureqa} and human baselines~\cite{figureqa} for different question templates. The results are from a subset of the Test 2 split in FigureQA. As mentioned in the main document, Test 2 split consists of chart images where the charts have alternated colors compared to the training set, such that the colors are novel for a given chart-type. Test 2 annotations are not publicly available and the results were obtained by sending model predictions to the authors. As seen in table \ref{tab:extra-results}, PReFIL outperforms RN for all question templates by a large margin and also outperforms human baseline in 12 out of 15 question templates.

\section{More Discussion of Example Outputs}

\begin{figure*}[!t]
    \centering
    \includegraphics[width=0.95\linewidth]{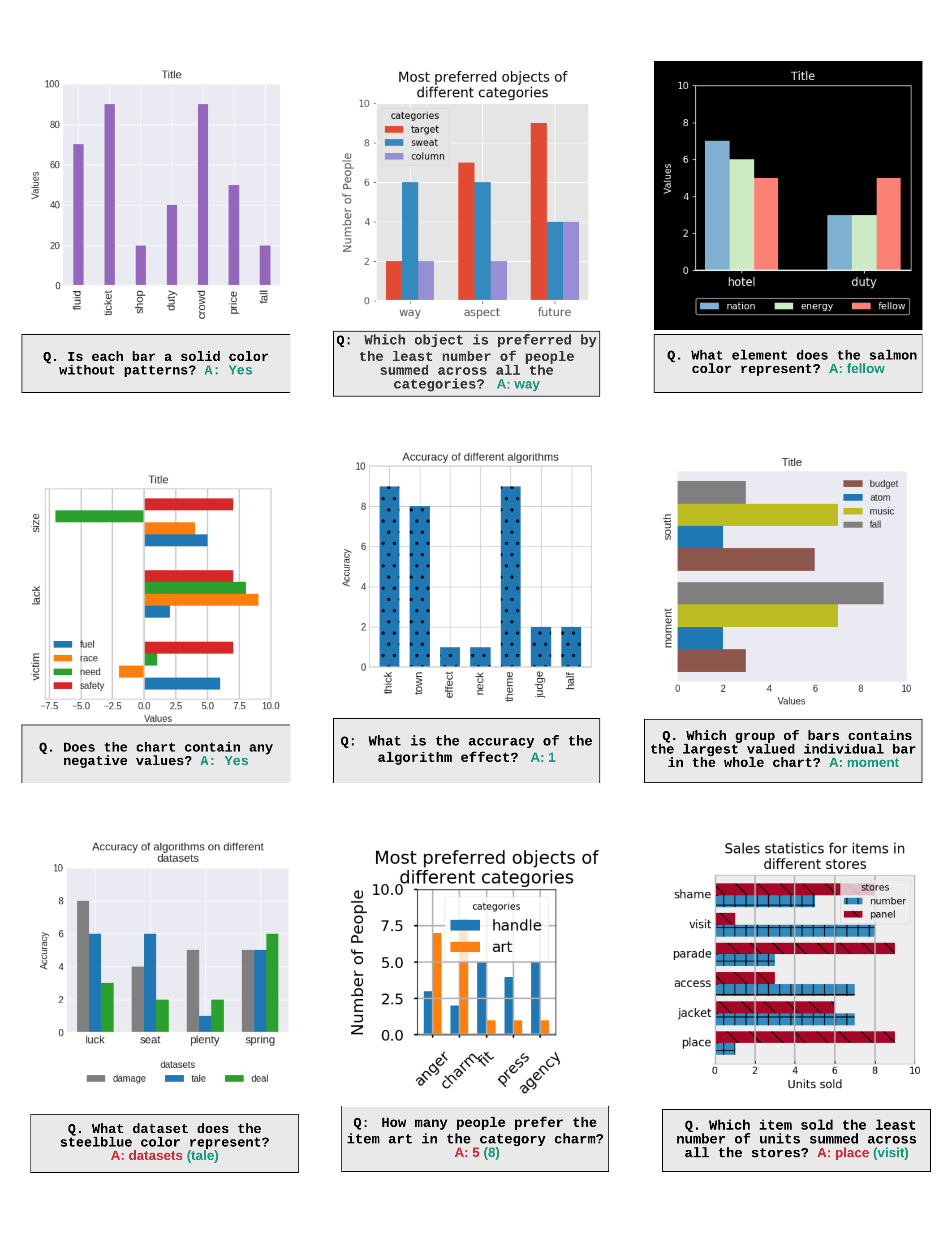}
    \caption{Some example predictions for PReFIL on the DVQA dataset. Red denotes incorrect predictions. For incorrect predictions, correct answer is shown in parenthesis.  \label{fig:dvqa-examples-supp}}
\end{figure*}

\begin{figure*}[!t]
    \centering
    \includegraphics[width=0.95\linewidth]{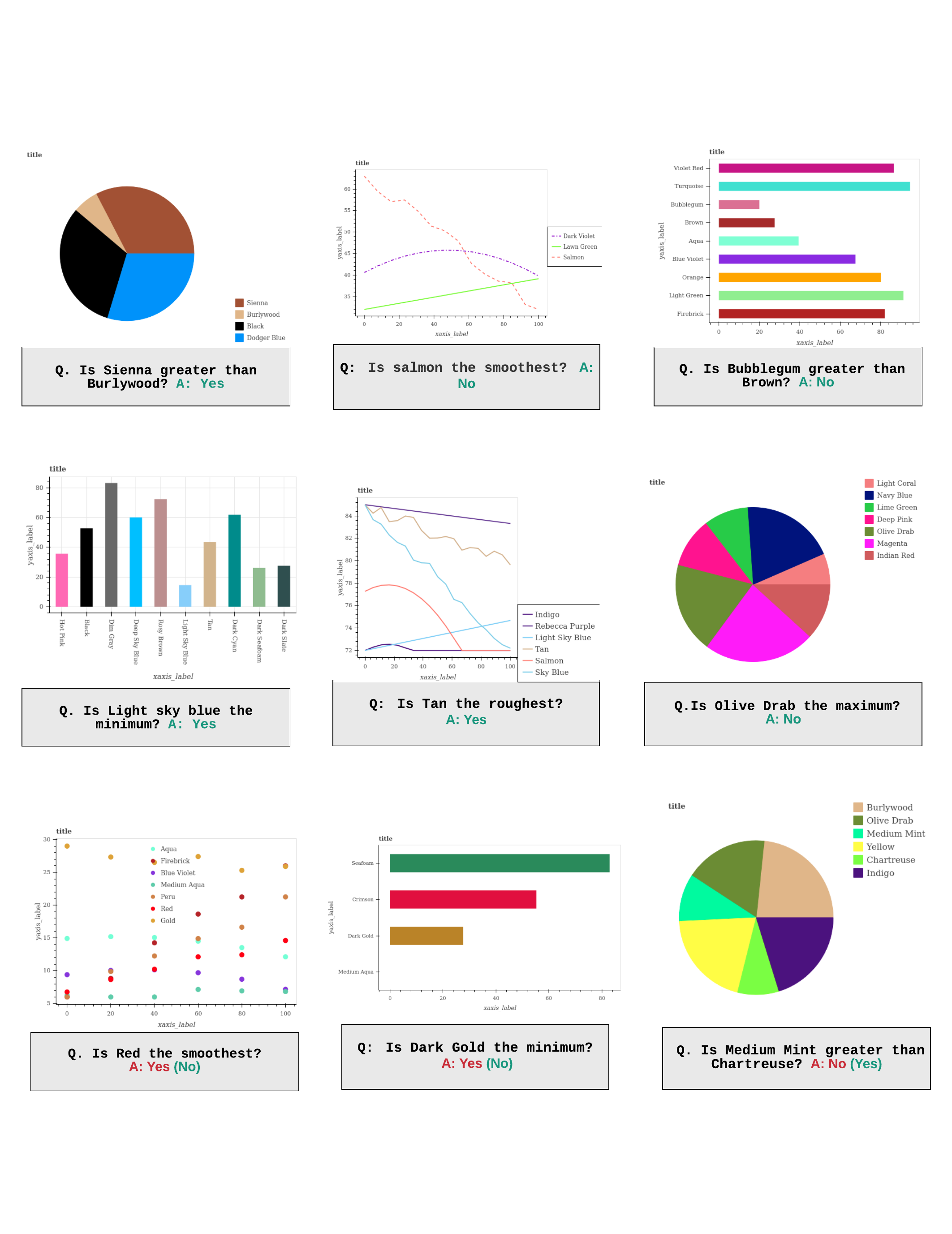}
    \caption{Some example predictions for PReFIL on the FigureQA dataset. Bottom row shows some incorrect predictions made by PReFIL. Red denotes incorrect predictions. For incorrect predictions, correct answer is shown in parenthesis.  \label{fig:figureqa-examples-supp}}
\end{figure*}

We present additional examples for our PReFIL algorithm for both the DVQA \cite{kafle2018dvqa} (Fig.~\ref{fig:dvqa-examples-supp}) and FigureQA (Fig.~\ref{fig:figureqa-examples-supp}) datasets. For both datasets, we present examples of correct predictions for a variety of examples (top two rows) and some cases of incorrect predictions (bottom row).

For DVQA, PReFIL with oracle OCR is exceedingly capable, with accuracy of over 96\% (see main text for details), but it makes some occasional errors. First, since the dynamic encoding is based on the position of words in the chart, PReFIL may detect the wrong word when the words are in close proximity to each other (Fig.~\ref{fig:dvqa-examples-supp}, bottom left). Second, when the chart elements are partially or fully obscured by the legend, PReFIL often fails to correctly parse the chart data (Fig.~\ref{fig:dvqa-examples-supp}, bottom center). Finally, for some charts, questions involving multiple measurements are also erroneous, especially when the measurements differ only by a small amount (Fig.~\ref{fig:dvqa-examples-supp}, bottom right).

For FigureQA, PReFIL again performs well across all categories, surpassing overall human accuracy. PReFIL is capable of answering a wide range of questions across several types of images (Fig. \ref{fig:figureqa-examples-supp}, top 2 rows). However, PReFIL often struggles for question template ``Is X the smoothest/roughest?'' especially for the dot-line style graphs. The errors are more prominent when the legend obscures or intermingles with the chart elements (Fig. \ref{fig:figureqa-examples-supp}, bottom left). Since the dots are not connected to each other, it is an extremely difficult task even for attentive human observers. Similarly, PReFIL makes occasional mistakes when comparing elements that are very close to each other (Fig. \ref{fig:figureqa-examples-supp}, bottom center and right). However, as seen in Table \ref{tab:extra-results}, PReFIL is more accurate than even human observers for comparing two elements.

\end{document}